\def\year{2020}\relax
\documentclass[letterpaper]{article} 
\usepackage{aaai19}  
\usepackage{times}  
\usepackage{helvet} 
\usepackage{courier}  
\usepackage[hyphens]{url}  
\usepackage{graphicx} 
\def\UrlFont{\rm}  
\usepackage{graphicx}  
\usepackage[dvipsnames]{xcolor}
\usepackage{tikz}
\usepackage{pgfplots}
\usepackage{appendix}

\makeatletter
\renewcommand\paragraph{\@startsection{paragraph}{4}{0pt}%
  {5pt}
  {-\parindent}
  {\textbf}}
\makeatother

\newcommand{\nop}[1]{}

\title{Automatic Discovery of Meme Genres with Diverse Appearances}

\author{William Theisen, Joel Brogan, Pamela Bilo Thomas, \\ 
\bf \Large Daniel Moreira, Pascal Phoa, Tim Weninger, Walter Scheirer\\
 Department of Computer Science and Engineering, University of Notre Dame
}

\begin{document}

\maketitle

\begin{abstract}
Forms of human communication are not static --- we expect some evolution in the way information is conveyed over time because of advances in technology. One example of this phenomenon is the image-based meme, which has emerged as a dominant form of political messaging in the past decade. While originally used to spread jokes on social media, memes are now having an outsized impact on public perception of world events, making them an important focus of study. 
A significant challenge in automatic meme analysis has been the development of a strategy to match memes from within a single genre when the appearances of the images vary greatly. In this paper we introduce a scalable automated visual recognition pipeline for discovering meme genres of diverse appearance. This pipeline can ingest meme images from a social network, apply computer vision-based techniques to extract local features and index new images into a database, and then organize the memes into related genres. To validate this approach, we perform a large case study on the 2019 Indonesian Presidential Election using a new dataset of over two million images collected from Twitter and Instagram, and examine a collection of humorous memes posted to Reddit. Results show that this approach can discover new meme genres with visually diverse images that share common stylistic elements, paving the way forward for further work in semantic analysis and content attribution. 
\end{abstract}

\section{Introduction}

\begin{figure}[t]
\centering
\includegraphics[width=0.48\textwidth]{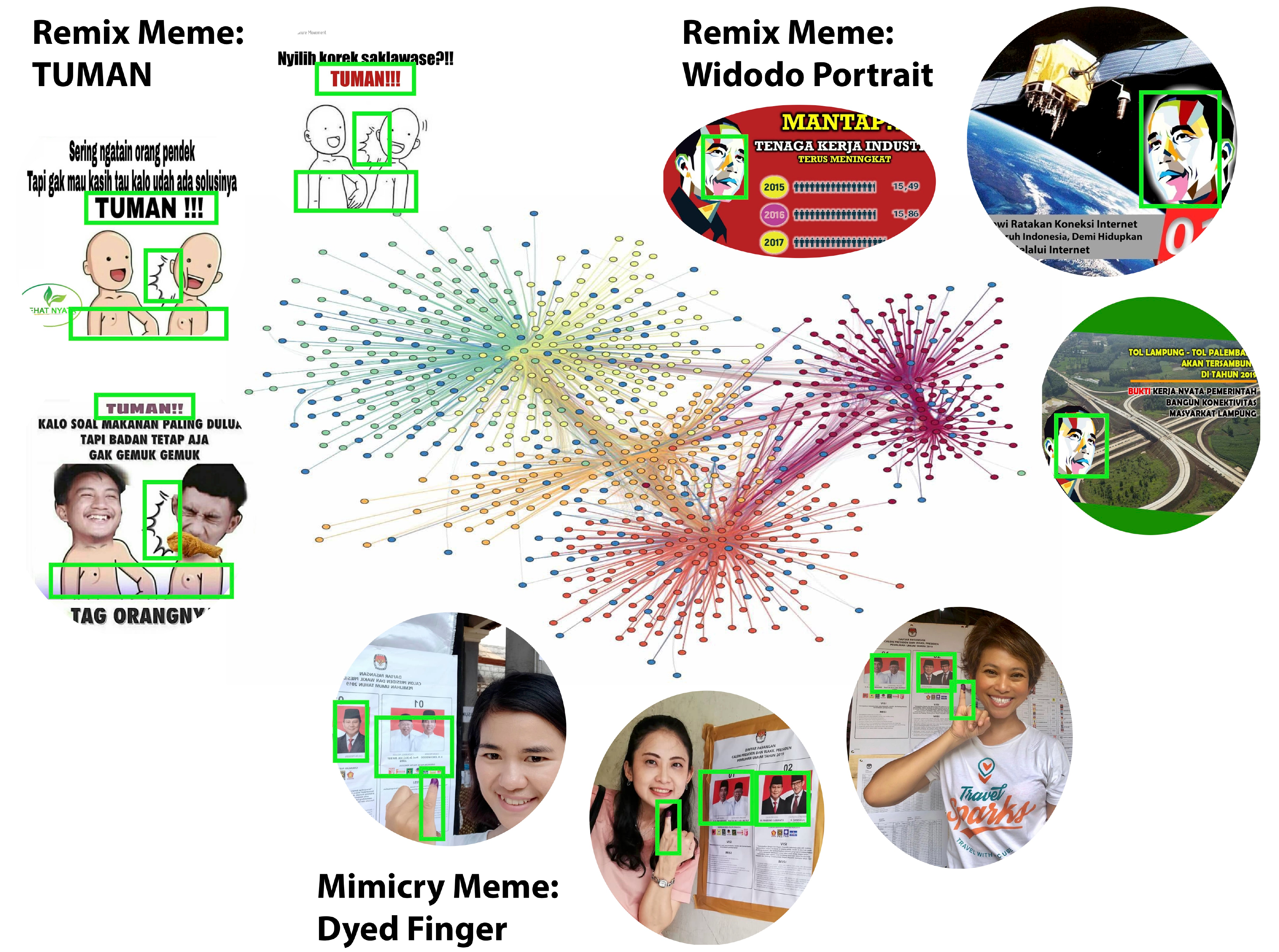} 
\caption{Individual meme genres can take on very different visual forms depending on the region and context in which they are found. In this work, we introduce an end-to-end processing pipeline that can ingest meme-style images from social media and automatically group them into meaningful genres based on small localized regions. The primary advantage of our approach is that matches (highlighted in green) are based on detected objects and themes, which are the building blocks of meme genres.
}
\label{fig:teaser}
\end{figure}

From silly cat photos to parodies of political candidates, image-based memes represent a vibrant and vital form of communication on social media. While once primarily the domain of the online communities found on 4chan and Reddit, most casual users of social media are now familiar with the concept of a meme and spread such content freely. Moreover, advances in image-editing software have made sophisticated tools accessible to untrained users, which has rapidly increased the production of memes by amateurs and professionals alike. When it comes to politics, the messages contained within memes span a landscape from ``get out the vote" campaigns~\cite{wsj} to anti-social narratives stoking violence~\cite{bbc,rumata2018net}. To monitor social media for emerging trends in memes, researchers have proposed new ways to use computer vision to automatically discover and track specific genres~\cite{BESKOW2020102170,dubey2018memesequencer,zannettou2018origins}. However, a significant challenge in this task is matching small, localized objects that may be the key to a particular image's membership in a genre (Fig.~\ref{fig:teaser}), where mimicry (\textit{i.e.}, reenacting a scene) or remix (\textit{i.e.}, manipulating an image) are common~\cite{shifman2014memes}. Towards the objective of improving social media analysis, this is the problem we address in this work.
 
To study this problem, we performed an expansive investigation of the 2019 Indonesian election, which was held on April 17, 2019, where a diverse set of memes and other political image content circulated on social media platforms. We collected a dataset of over two million related images from social media, including over 174,000 images from sources on Twitter, and almost 1.9 million images from sources on Instagram. The images were taken from the time period covering May 31, 2018 to May 31, 2019. Our primary goal was to detect and study the election-related trends that were expressed in images and shared on social media. Our investigation found spontaneous and coordinated campaigns to influence different aspects of the election, which are introduced and analyzed in the present work. 

\begin{figure}[t]
    \centering
    \includegraphics[width=0.38\textwidth]{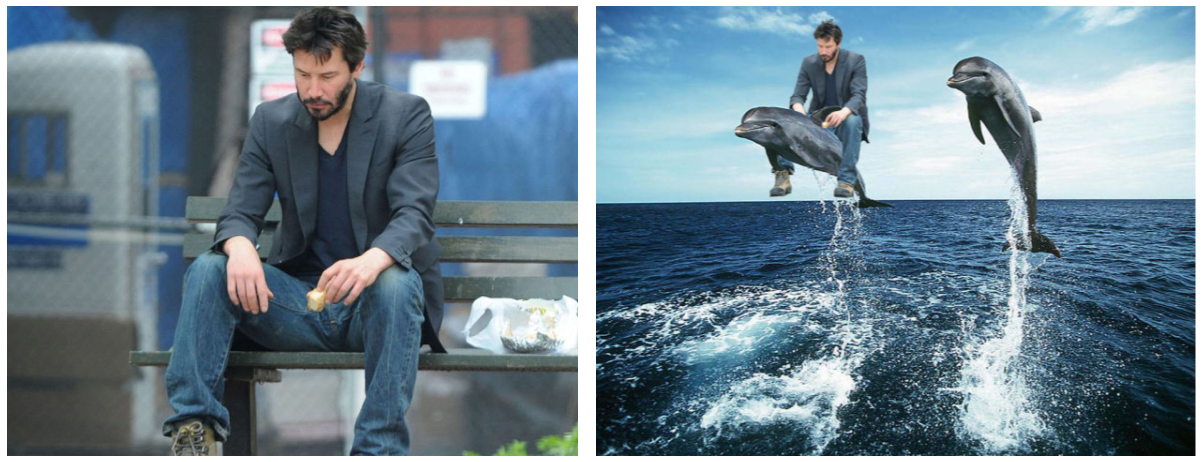}
    \vspace{-3mm}
    \caption{Two instances of the ``Sad Keanu" meme that share a local object (the actor Keanu Reeves), but have significantly different global appearances. This example is taken from the Reddit Photoshop Battles meme dataset~\protect\cite{moreira2018image}, which we experiment with in this paper.}
    \label{fig:sad}
    \vspace{-3mm}
\end{figure}

In support of this effort, we developed a general visual recognition pipeline that can discover meme genres in images of diverse appearance from social media.
A number of meme clustering approaches have been suggested in recent literature~\cite{dang2015visual,dubey2018memesequencer,zannettou2018origins}, which are effective in situations where only small changes have been made to the images during remixing, or when meta-data is available. In contrast, our pipeline relies purely on visual features extracted from localized objects, and is designed to match both mimicry (\textit{e.g.}, the hand gestures shown in Fig.~\ref{fig:teaser}) and remix (\textit{e.g.}, the variants of two illustrated characters shown in Fig.~\ref{fig:teaser}). {A more specific (and perhaps familiar) example of meme images that have shared local features, but a different global appearance can be seen in Fig.~\ref{fig:sad}. Approaches relying on global feature representations struggle on such pairs of images, which is what we aim in part to address in this work.} 

Indonesia was chosen as a primary case study for a variety of reasons, including its large number of Internet users, diverse population, and democratic culture. 
As a large, middle-income democracy~\cite{WorldBank}, Indonesia provided a unique opportunity to study the spread of political memes during the 2019 election in a population that is relatively new to the Internet. Results gathered from this case study could be applied to model what might happen in other middle-income democracies with recent Internet participation. Further, opinion polls and election results allowed us to monitor the views the Indonesian people held about their government, and who would ultimately win the election, as a result of, or in spite of, the memes that were spread beforehand and the information the public was exposed to. The findings presented in this paper stem from a new mode of inquiry: a large-scale visual analysis of political meme content associated with a single major world event.

To summarize, our contributions in this paper are:
\begin{itemize}
    \item A new dataset of over two million visually diverse images related to the 2019 Indonesian presidential election, collected from Twitter and Instagram. These images constitute one of the largest open access collections related to a single political event.
    \item An end-to-end visual recognition pipeline for meme genre discovery that is able to operate even when the images within a genre do not have a strong global style. The pipeline has been designed to scale to process images on the order of millions of instances.  
    \item Qualitative and quantitative analyses of the proposed pipeline, using the data from the 2019 Indonesian election as a case study, which demonstrate the viability of the proposed approach and highlight new political findings for the event. {An analysis is also performed on the Reddit Photoshop Battles meme dataset~\cite{moreira2018image} to show that the approach generalizes to other settings.}  
    \item A set of recommendations for research related to the automatic visual analysis of Internet memes, including the technical and social aspects.
\end{itemize}

\section{Related Work}
\paragraph{Meme genres.}

\begin{figure}[t]
    \centering
    \includegraphics[width=0.47\textwidth]{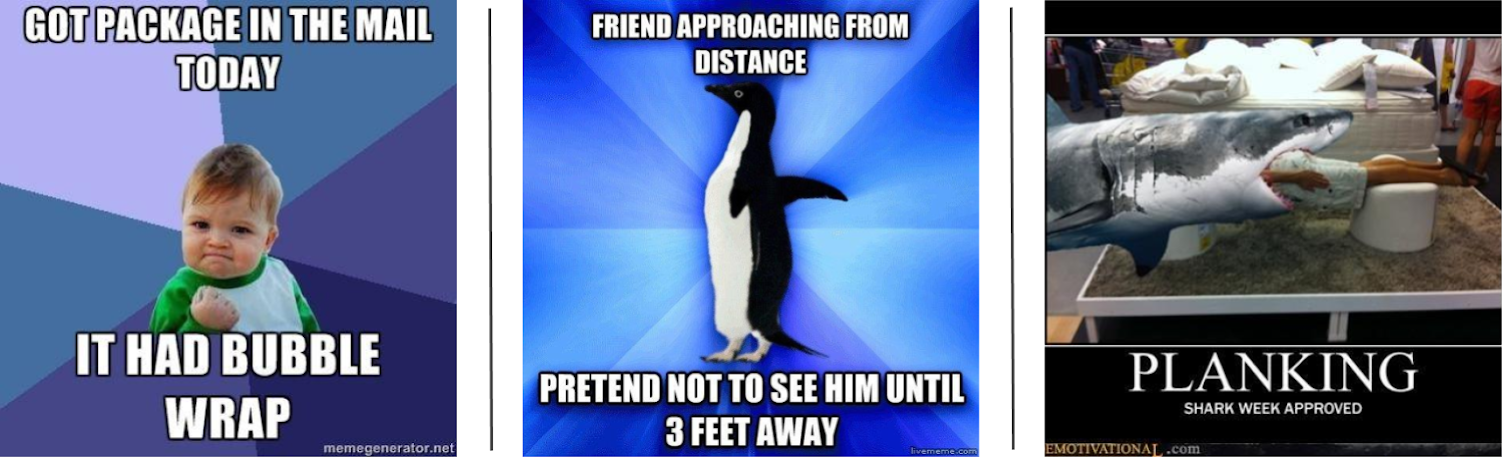}
    \vspace{-3mm}
    \caption{Examples of the  meme genres ``Success Kid," ``Socially Awkward Penguin," and ``Planking."}
    \label{fig:popmemes}
    \vspace{-3mm}
\end{figure}

Shifman described the concept of a \emph{meme genre} as being a diverse form categorization~\cite{shifman2014memes}. Genres range from near-duplicate memes with a well-defined content style, such as Stock Character Macros (\textit{e.g.}, ``Success Kid,'' ``Socially Awkward Penguin''), to semantically similar memes whose content is much more unpredictable, such as Reaction Photoshops (\textit{e.g.}, Reddit's r/photoshopbattles), and Photo Fads (\textit{e.g.}, silly activities like ``planking"). Some examples of these genres are shown in Figs.~\ref{fig:sad} and~\ref{fig:popmemes}.
These genres highlight meme variety, implying that automated analysis must address visual diversity.

\paragraph{Non-automated meme analysis.}
One important vein of the related research is the qualitative analyses of memes.
Some researchers target a particular social media ecosystem in search of relevant memes, such as YouTube \cite{shifman2011anatomy}, the Reddit Quickmeme board \cite{coscia2013competition}, and 4chan's random board /b/ \cite{nissenbaum2017internet}. Others track a specific meme of interest, such as the ``It Gets Better'' viral media campaign \cite{gal2016gets}, and the ``Obama Hope'' poster \cite{seiffert2018memes}.
Although these studies were conducted via traditional research methods, they constitute an important step towards understanding the process of drawing inferences from genres of memes.

\paragraph{Meme clustering.}
Good initial progress has been made on automatically tracking and clustering memes.
Dang et al. introduced a strategy to cluster memes posted on Reddit by relying on the textual content of posts (\textit{e.g.}, accompanying hashtags, comments, URLs), as well as the textual content of websites hosting near-duplicate versions of the memes found through Google reverse image search \cite{dang2015visual}.
Different from our proposed algorithm, their idea is to use text clustering as a proxy for meme  clustering, instead of using visual content to derive clusters.

Dubney et al. focused on the meme genre of Stock Character Macros by developing image processing methods to recover the meme's template image (such as the ``Success Kid'') and segment the meme text for further application of OCR \cite{dubey2018memesequencer}.
Deep learning-based neural networks are used to extract both visual and textual features, which are concatenated to generate a single meme representation for clustering. Beskow et al. describe a similar approach underpinned by deep learning for feature extraction \cite{BESKOW2020102170}. In both cases, features are extracted in a global manner. Consequently, this type of strategy is only able to match images presenting either near-duplicate visual or textual content. Our algorithm does not have this constraint.

Zannettou et al. propose an image processing pipeline to group and track memes that relies on perceptual hashing (PHASH) to describe and compare visual content \cite{zannettou2018origins}.
In addition, their pipeline requires a set of previously annotated memes, such as the ones archived by the website KnowYourMeme\footnote{\url{https://knowyourmeme.com/}}, to define templates that are used to drive the data clustering process. In contrast, our approach does not need previous knowledge about the analyzed memes. This is an important attribute that allows our pipeline to work on data from a case like the 2019 Indonesian election, where the content might not receive enough global attention to be archived by a meme aggregator. 

\paragraph{Content-based image retrieval.}
The proposed pipeline is related to the field of Content-Based Image Retrieval (CBIR).
In general, CBIR algorithms focus on two different matching tasks.
The first matching task aims to retrieve near-duplicate images with content depicting a specific element of interest under limited variation.
This is the case of tasks such as object tracking, \textit{e.g.}, track a specific vehicle plate across the frames of a street video~\cite{cehovin2016visual}, place recognition, \textit{e.g.}, retrieve images of the Eiffel Tower~\cite{lowry2016visual}, and strict near-duplicate detection~\cite{winkler2013nd,lowry2016visual}.
The second matching task, by contrast, aims to retrieve images whose contents are semantically similar.
This is the case of tasks such as object recognition~\cite{russakovsky2015imagenet}, \textit{e.g.}, retrieve images containing cats of any breed or color, and scene recognition, \textit{e.g.}, retrieve images of different golf courses~\cite{zhou2017places}. Both tasks can be used to retrieve different genres of memes. 

In many CBIR searches involving memes, users may not actually known what to expect in response to a given query.
Take, for example, images containing small spliced objects, people in similar poses, and subtly altered scenes.
Should the results include near-duplicates or semantically similar images?
Should they match on the foreground or the background?
The task of meme analysis requires the retrieval of every meaningful piece of the query in order to provide satisfactory material for data clustering.
To cope with this problem, Brogan et al. introduce an algorithm for the problem of matching small objects between two different images \cite{brogandynamic}. Called the Objects-in-Scene To Objects-in-Scene (OS2OS) algorithm, this approach allows for retrieval of related images that vary greatly in appearance, such as the hand gestures depicted in the images on the bottom of Fig.~\ref{fig:teaser}.
In this work, we use the OS2OS technique as a basis for the image comparison step of the proposed pipeline.

\paragraph{Image forensics.}
The present work is also related to image forensics.
Algorithms for the detection of inconsistencies in images --- \cite{chen2019secure,huh2018fighting,farid2016photo} --- do not help for the analysis of memes because most of the instances are openly manipulated. That is, the fact the image is somehow altered (with added text, etc.) is not hidden. However, algorithms related to the semantic analysis of images --- such as image phylogeny~\cite{oliveira2016multiple,dias2013large} and image provenance analysis~\cite{bharati2019beyond,moreira2018image} --- are closer to meme discovery.
The difference is in the final output and its purpose. Beyond telling us that manipulation has occurred, forensics tools can help us understand and track how meme genres change over time as content is edited.

\paragraph{Content understanding.}
Finally, this work can aid further research in the broader field of social media content understanding, in the spirit of work such as Dewen et al. \cite{dewan2017towards} and Blandfort et al. \cite{blandfort2019multimodal}.

\section{Meme Genre Discovery}
\begin{figure*}
\centering
\includegraphics[width=0.9\textwidth]{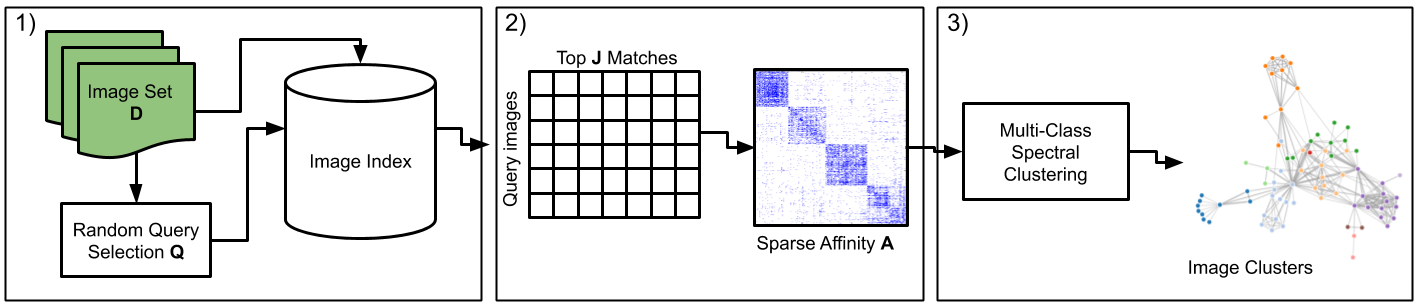} 
\caption{Overview of meme genre discovery system. Given a set of images, the pipeline will create genre clusters based on an image index built from local features. These local features allow for matching based on local regions shared between images.
\vspace{-3mm}
}
\label{fig:pipeline}
\end{figure*}

To discover meme genres in large collections of images, we developed a new multi-stage pipeline that detects related images based on the visual content they share. In contrast to prior work, our Meme Genre Discovery (MGD) pipeline attempts to establish subtle content relationships based on localized similarities between images (see Fig.~\ref{fig:teaser}). It is designed to construct, in an unsupervised way, a clustering of content within the dataset --- without any human intervention or \textit{a priori} knowledge. To accomplish this, the MGD pipeline, shown in Fig.~\ref{fig:pipeline}, has three core steps: 

\begin{enumerate}
    \item[] \textbf{Step 1.} Indexing of images based on local features. 
    \item[] \textbf{Step 2.} Construction of a graph that expresses relationships between images, represented by an affinity matrix.
    \item[] \textbf{Step 3.} Spectral clustering of the graph to establish meme genres of related images.
\end{enumerate}

\noindent The remainder of this section describes each step in detail.


\paragraph{Step 1: Image indexing.}
To process data on the order of millions of images, we require an image index to facilitate fast and scalable searches.  The performance of a given image index and search strategy depends on the nature of the image feature representation, indexing structure, and matching similarity metric chosen. For these purposes we implemented a version of the indexing strategy from the OS2OS algorithm~\cite{brogandynamic}, which allows us to provide a query image to the index, and match images that have either a large global similarity or many local matches to the query.

Image features describe an aspect of a target spatial region of an image. They are, ideally, invariant to various image transformations (\textit{e.g.}, rotation, translation, scale) to improve matching, and discard data that is not useful to the matching process. The index here (Fig.~\ref{fig:pipeline}.1) is built using features extracted from localized patches within individual images and provides image match results based on local regions shared between images. This allows us to match in either a local or global context, depending on the number of correspondences found between images. The specific feature representation used is a 64-dimensional SURF feature~\cite{bay2006surf}, which is a compact representation designed for very fast matching. It remains much faster than other viable feature representations, including features from deep learning approaches~\cite{moreira2018image}. The index used with these features is an Inverted File (IVF) index with 2,048 8-dimensional centroids trained via Optimized Product Quantization (OPQ)~\cite{ge2013optimized}. We extract and compress 2,500 features for all $N$ images in a dataset $D$ to submit into the IVF. 

\paragraph{Step 2: Affinity matrix.}
After the index is built, we use it to construct an approximate affinity matrix $A$ of size $N \times N$ (Fig.~\ref{fig:pipeline}.2) that characterizes the relationships between images in the index as a graph. To do this, a set of images $Q$ is randomly sampled from the dataset $D$ and is used to query into the index. Other strategies for querying exist, such as selecting query images by identifying images that have been modified in some way using media forensics tools~\cite{farid2016photo}. However, for our experiments described below, random querying was sufficient. If $D$ consists of a collection of unordered memes, then $Q$ will be meme instances that will match other instances from the same genre in the index if they exist. For each query $q_i \in Q$ we collect a set of $J$ total match results $M_{q}$ from the index. Each match $m_j \in M_q$ is then considered a weighted edge such that $A_{q_i,m_j}=S(q_i,m_j), j \le J$, where $S(q,m)$ is the OS2OS affinity (\textit{i.e.}, feature correspondence) score~\cite{brogandynamic} between query image $q$ and matched image $m$. 

{$|Q|$ and $J$ scale proportionally with the number of edges in the matrix. $|Q|$ scales with respect to the number of nodes with edges, $J$ increases alongside node ranks. Our spectral clustering method utilizes multi-degree edge paths to compute an embedded distance space. Therefore, we determined a minimally connected (or nearly connected) graph would be sufficient to perform the clustering while keeping computational overhead low. In practice, we set both $|Q|$ and $J$ based on small-scale experiments on the minimum parameters required to construct a connected graph. We determined that a sample size of $|Q|=0.1\times N$, $N$ being the dataset size, will reliably provide a connected graph when $J=100$.} 


\paragraph{Step 3: Spectral clustering.}
After building the affinity matrix $A$, we perform multiclass spectral embedding and clustering~\cite{stella2003multiclass} to assign each image to a hypothesized meme genre. The relationship data for all $N$ images is already in an approximate affinity matrix format, allowing us to employ the spectral clustering method by diagonalizing and then decomposing $A$ into its principal components via eigendecomposition. The  $N$-dimensional eigenvectors $V$ computed by this decomposition define the $|V|$ axes along which unique clusters lie. By utilizing these vectors as columns in a new $N \times V$ Matrix $R$ we obtain a re-projected set of coordinates for all $N$ images in the dataset $D$. The spectral clustering of $A$ takes advantage of relationships between images spanning multiple edges (\textit{e.g.}, image $I$ and $P$ both share a visual object with image $R$, meaning $I$ and $P$ are related). This non-linear re-projection allows sets of images with subtle yet meaningful relationships to reside closely together in a content-sharing space, even if those clusters were originally highly amorphous and non-convex. K-means clustering is run on $V$ to obtain the final clusters that ideally represent the meme genres (Fig.~\ref{fig:pipeline}.3). {In this paper, we use the term cluster to specifically refer to the output of the K-means algorithm.} 

\begin{figure*}[t]
    \centering
    \includegraphics[width=1.0\textwidth]{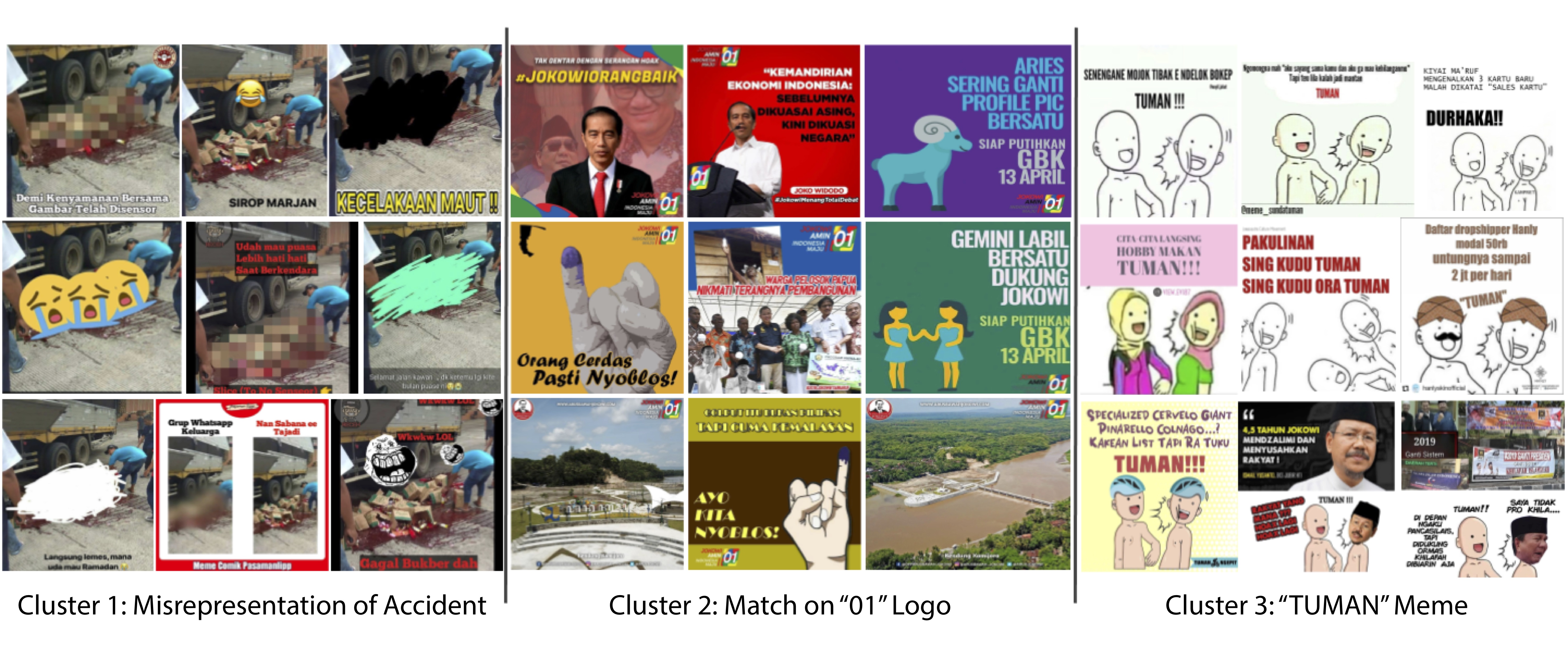}
   \vspace{-8mm}
    \caption{Example images from three of the genres detected by our approach for the entire 2019 Indonesian national election dataset. The first set shows a genre that contains images claiming a truck had run over a man. However, also in the genre was a fact checking image showing this was not the case (bottom center). The second set was matched on the small ``01'' logo in all of the images, indicating support for candidate Widodo (the 1st candidate on the ballot); this example highlights our approach's localized matching capability. The third set is an example of an Indonesian meme genre that looks similar to those commonly found in the West. This particular genre makes a joke about an action that will likely repeat, even if it is foolish. The word ``Tuman," loosely translated, means ``basic". As in, it is a basic fact that something is likely to repeat.} 
    \label{fig:clusterExamples}
    \vspace{-3mm}
\end{figure*}

\section{Experiments}

To evaluate the utility of the MGD pipeline, both data and validation procedures must be considered.
With respect to data, we required a dataset of images drawn from a real-world event with a great deal of variation in visual appearance. The goals of our experimental evaluation were then to (1) detect useful clusters from millions of images; (2) verify that the images contained within a cluster are meaningful to human observers;  (3) compare the performance of the MGD pipeline with existing global feature matching approaches~\cite{zannettou2018origins,dubey2018memesequencer}; {(4) verify generalization ability across datasets; and (5) assess sensitivity to the specified number of clusters.} 

\paragraph{Data collection.}
We collected a large-scale dataset containing meme-style imagery and other political images that circulated on social media in the months leading up to the 2019 Indonesian national election. This election, which occurred on April 17, 2019, was a rematch between the two candidates that competed in the 2014 election. The incumbent, Joko ``Jokowi" Widodo, the center-left candidate who appealed to younger voters as a ``man of the people," declared victory over the militaristic, strongly Islamic candidate, Prabowo Subianto, with over 55 percent of the vote~\cite{cnn}. After the election, the losing party alleged fraud and massive protests shortly followed, during which at least six people died and 200 were injured. As part of the government's response, social media sites were temporarily blocked nationwide~\cite{bbc}. Much of the drama unfolding during these deadly events was promoted and captured on social media. Under this backdrop we deployed our meme analysis pipeline.

We began collecting images on May 31st, 2018, (11 months preceding the Indonesian presidential election) and completed our collection on May 31st, 2019 (6 weeks following the election). We gathered data from Twitter and Instagram. {The images from Twitter were collected using a Google Chrome plugin\footnote{https://chrome.google.com/webstore/detail/twitter-media-downloader/cblpjenafgeohmnjknfhpdbdljfkndig?hl=en}. They were collected from 11 hashtags and 3 users. Images collected from Instagram were collected by following 15 hashtags and 5 users. They were downloaded with a Python program\footnote{https://github.com/althonos/InstaLooter}.} In total we collected 174,328 images from Twitter and 1,851,411 images from Instagram. Hashtags and users that were sources are listed in the Appendix. Sources on both sites were identified with the help of our partners at CekFakta\footnote{https://cekfakta.com/}, an Indonesian fact checking organization. They are sponsored by the Google News Initiative and have working partnerships with most of the large news outlets in Indonesia. CekFakta monitors social media streams in Indonesia for misinformation. They also have a reporting feature on their website so that the public may  contribute to the process. All collected data was harvested from public sources and was meant to be shared.

\paragraph{Genre detection and validation. } 
We used the MGD pipeline to discover meme genres using the union of images collected from Twitter and Instagram. In total MGD reported 7,691 clusters; Fig.~\ref{fig:clusterExamples} shows three example genres in three separate clusters with 9 exemplar-images each. With respect to the computational efficiency of this approach, runtime scales linearly with the number of images and indexing can be performed in parallel. Approximately 10,000 images can be processed per hour.

\begin{figure}[t]
    \centering
    \includegraphics[width=0.45\textwidth]{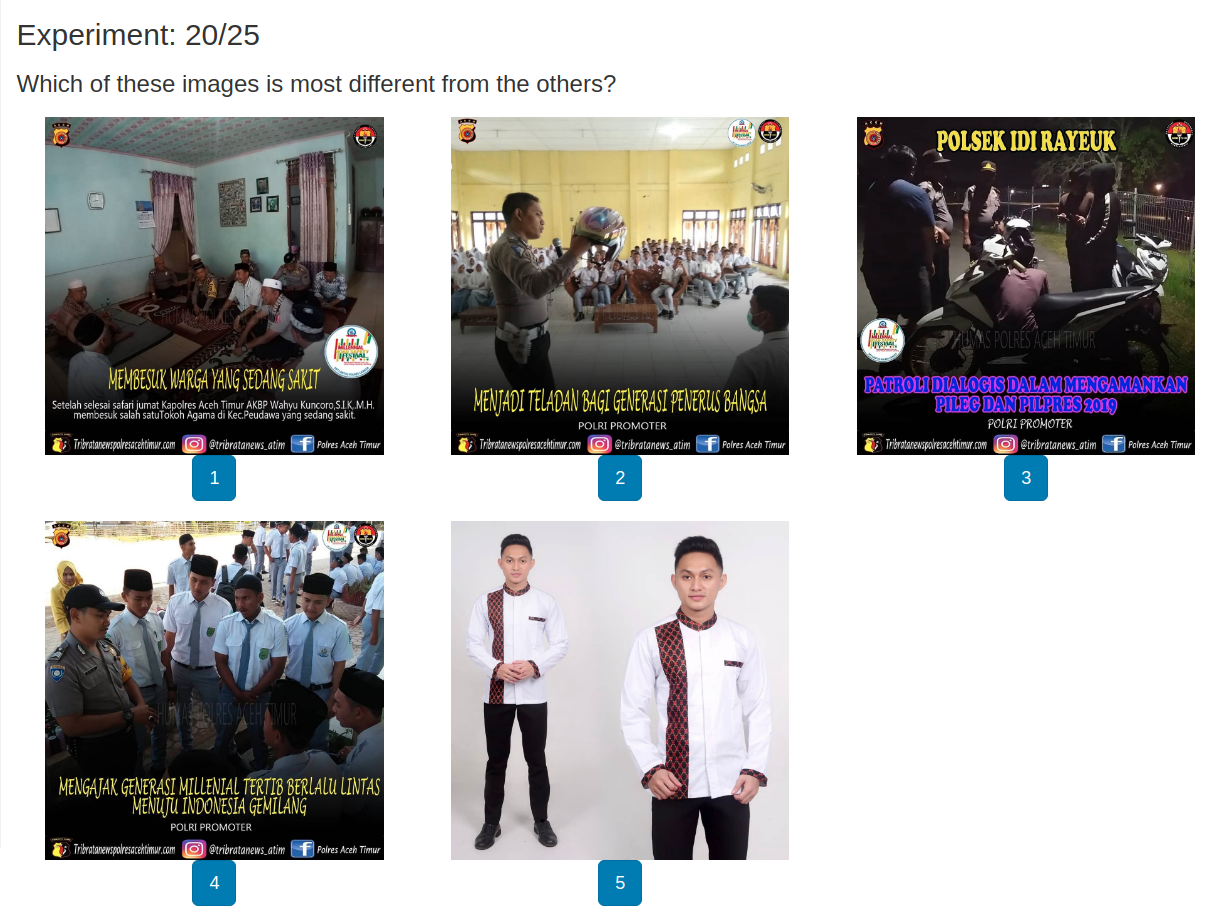}
    \caption{An example of what a single question in the impostor-host experiment looks like to a Mechanical Turk worker. The goal of the worker is to pick out the image that does not belong in the cluster (in this case, image 5).}
    \label{fig:experimentExample}
    \vspace{-3mm}
\end{figure}

Evaluating the validity of the MGD pipeline and resulting clusters requires judgement of the visual coherence of the images within a resulting cluster. As in cluster analysis, this type of evaluation involves judgement of the intra-cluster similarity. Unfortunately, manually labelling two million images to establish ground-truth data is not feasible. Instead, we validated the coherence and interpretability of the detected clusters using a straightforward human annotation process. 
Specifically, we use the impostor-host methodology, which is  used to evaluate clustering algorithms in the absence of a ground-truth clustering~\cite{10.1145/2396761.2396843}. In this methodology, human annotators are presented with five images: four from the same ``host'' cluster and a single ``impostor'' image, which is randomly selected from another cluster. The annotator is then asked to select the impostor. An example of this task is shown in Fig.~\ref{fig:experimentExample}. The motivation for this is that the more closely related the images in a cluster are, the easier it should be for a human observer to correctly identify the impostor image. 


We used Amazon's Mechanical Turk service to recruit 1,869 human annotators. Each annotator was paid 0.25 USD to annotate 25 tasks each. Of these, five were simple pre-labeled control tasks designed to ensure that workers provided honest labels. If a worker missed two or more of the control tasks then their annotations are discarded. In total we discarded the annotations from 31 annotators resulting in 1,838 qualified human annotators who contributed 24,060 total annotations.
An important consideration is that the overall population of human annotators was unfamiliar with Indonesian culture and politics, and therefore unlikely to possess any preconceived notions or context about the image content. Due to this disconnect, we are confident that the annotators relied almost entirely on visual cues.

The accuracy of randomly guessing the impostor image is 1/5 = 20\%. The overall accuracy of the human annotators was 51.21\%, significantly better than chance. 
These results lead us to believe that the MGD pipeline can detect coherent image clusters reliably. In other words, the impostor-host methodology demonstrated that a majority of the detected clusters had a cohesive-enough theme that was identifiable even in the presence of an impostor image. But further analysis was needed to understand the content in each cluster.

\paragraph{Human interpretation of generated clusters.}

\begin{figure}
\centering
\includegraphics[width=1.0\linewidth]{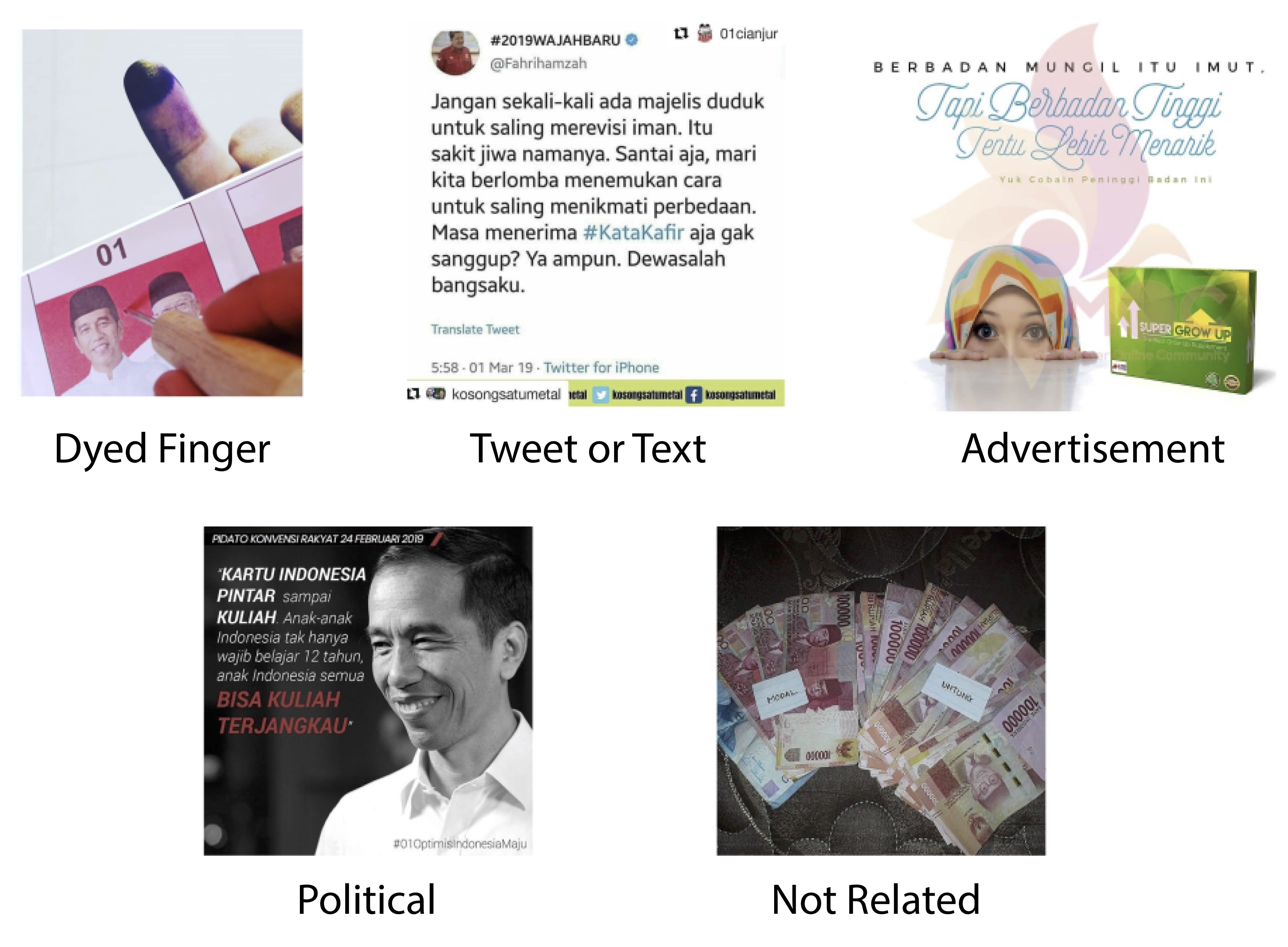}
\vspace{-4mm}
\caption{Representative examples from each of the five super-genre categories used as labels by the annotators of the clusters from the Indonesian national election dataset.}
\label{fig:categoryExamples}
\vspace{-3mm}
\end{figure}

After ensuring that the generated clusters were not just the product of chance, we recruited a team of undergraduate students to manually label them in order to get a sense of what kind of content they contained. We defined several ``super-genres" of content to serve as the labels: ``Advertisement,'' ``Dyed Finger,'' ``Tweet or Text,'' ``Political,'' and ``Not Related''. The latter super-genre was given to clusters where there was no obvious theme to the content, allowing us to quantify the fraction of clusters that are not useful. Each super-genre contained further sub-genres, other political content and content not directly related to the Indonesian election. A breakdown of the data set by super-genre can be found in Fig.~\ref{manual}. An example of each super-genre can be found in Fig.~\ref{fig:categoryExamples}.  

The super-genres were decided upon collectively by  the authors after reviewing an initial subset of labeled clusters generated by the undergraduate students. In this initial subset, the clusters were freely labeled by the undergraduate students. The most prevalent labels influenced the choice of the super-genre labels. The selection of super-genres was made in order to allow for uniformity across the labelling of the entire Indonesian national election dataset, and the five selected super-genres were decided upon to allow for a concise view of the data landscape.

While most of the super-genres we defined are very broad, there is one well-defined genre that our annotators identified: the ``dyed finger" mimicry meme. After individuals voted in the 2019 Indonesian election, they dipped their finger in ink to indicate that they had voted, which ensured that they could only vote once (a common practice in the developing world). On election day, many people posted images of their dyed fingers to social media in a ``get out the vote" campaign. Examples are shown in Figs.~\ref{fig:teaser} and~\ref{fig:categoryExamples}. 

\newenvironment{customlegend}[1][]{%
    \begingroup
    \csname pgfplots@init@cleared@structures\endcsname
    \pgfplotsset{#1}%
}{%
    \csname pgfplots@createlegend\endcsname
    \endgroup
}%

\def\addlegendimage{\csname pgfplots@addlegendimage\endcsname}

\begin{figure}
\def\angle{0}
\def\radius{\columnwidth/2}
\def\cyclelist{{"orange","blue","red","green","violet"}}
\newcount\cyclecount \cyclecount=-1
\newcount\ind \ind=-1
\begin{minipage}{.3\columnwidth}
\begin{tikzpicture}[nodes = {align=center}]
  \foreach \percent/\name in {
      9.49/Dyed finger,
      24.67/Not related,
      21.50/Advertisement,
      26.76/Political,
      17.58/Tweet or text
    } {
      \ifx\percent\empty\else               
        \global\advance\cyclecount by 1     
        \global\advance\ind by 1            
        \fi
        \pgfmathparse{\cyclelist[\the\ind]} 
        \edef\color{\pgfmathresult}         
        \draw[fill={\color!50},draw={\color}] (0,0) -- (\angle:\radius)
          arc (\angle:\angle+\percent*3.6:\radius) -- cycle;
        \node at (\angle+0.5*\percent*3.6:0.65*\radius) 
        {\percent};
        \pgfmathparse{\angle+\percent*3.6}  
        \xdef\angle{\pgfmathresult}         
    };
    \end{tikzpicture}
\end{minipage}
~{}\qquad{}~
\begin{minipage}{.35\columnwidth}
\begin{tikzpicture}[nodes = {align=center}]
    \begin{axis}[
            ybar,
            ymin=0,
            yticklabel pos=right,
            symbolic x coords={1,2,3,4,5},
            xticklabel=\empty,
            width=5.5cm,
        every axis plot/.append style={
          ybar,
          bar shift=0pt,
          fill
        }
        ]
      \addplot[green!50]coordinates{(1,2058)};
      \addplot[red!50]coordinates{(2,1653)};
      \addplot[violet!50]coordinates{(3,1352)};
      \addplot[orange!50]coordinates{(4,730)};
      \addplot[blue!50]coordinates{(5,1897)};
    \end{axis}
\end{tikzpicture}
\end{minipage}

\begin{tikzpicture}
\centering
\begin{customlegend}[legend cell align=left, 
legend columns=3,
legend entries={ 
Political,
Advertisement,
Tweet or Text,
Dyed Finger,
Not Related
},
legend style={
draw=none,
nodes={scale=0.75, transform shape},
column sep=1ex,}] 
    \addlegendimage{area legend,green,fill=green!20}
    \addlegendimage{area legend,red,fill=red!20}
    \addlegendimage{area legend,violet,fill=violet!20}
    \addlegendimage{area legend,orange,fill=orange!20}
    \addlegendimage{area legend,blue,fill=blue!20}
\end{customlegend}
\end{tikzpicture}

\caption{(Left) A breakdown by percentage of five meme super-genres as they appear in the clusters generated by our approach after it was applied over the complete dataset from the Indonesian national election. (Right) The corresponding numbers of clusters manually labelled by our annotators.}
\label{manual}
\vspace{-5mm}
\end{figure}
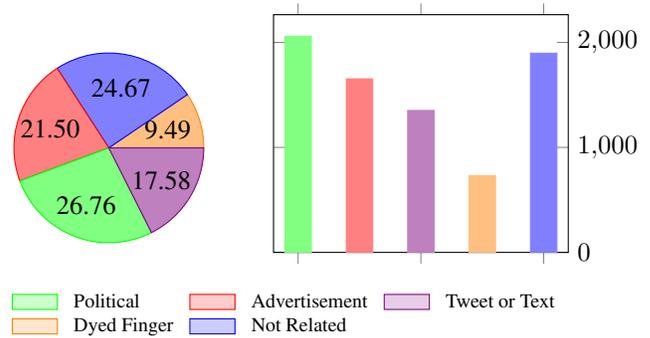


The dyed finger meme is a subset of political memes, but its clusters were so common in our dataset (nearly 10\% of all images considered) that we decided to classify it as its own genre. Given the negative sentiment surrounding political memes, such an overwhelming celebration of participatory democracy was an unexpected and encouraging finding of this work. The ability for our pipeline to discover the dyed finger meme, which is not a common meme in any Western dataset, speaks to the versatility of our algorithm in finding emerging memes.


\begin{figure*}[!ht]
\centering
\includegraphics[width=1.0\textwidth]{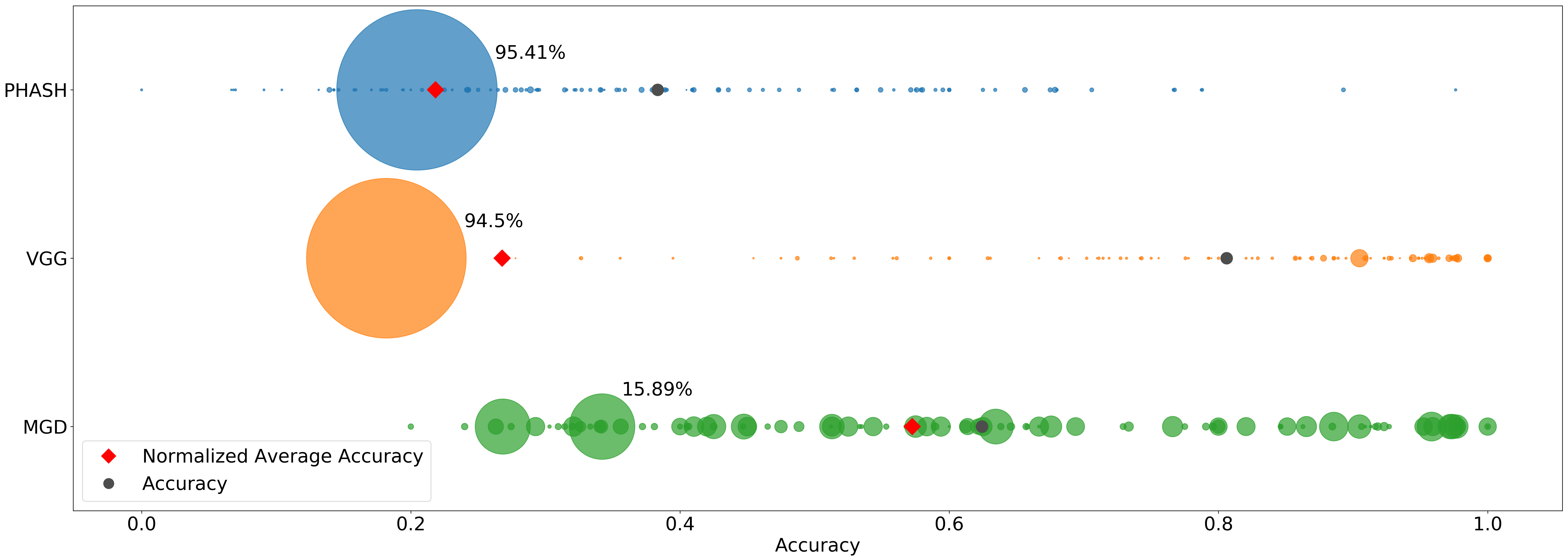} 
\caption{Results of the feature extractor comparison experiment on the smaller dataset of 44,612 images taken from the larger Indonesian national election dataset, organized graphically. Each circle is a cluster with a size proportional to the number of images in that cluster. They are arranged along the x-axis with respect to the accuracy rating for that cluster on the impostor-host task. The largest cluster of each method is labelled with the percentage of total images it contains. The most significant finding from this experiment is that the PHASH and VGG global feature extraction and matching methods put nearly all of the images in the dataset into a single cluster, and are not suitable for use with a dataset containing related images of diverse visual appearance. {Both Normalized Average Accuracy and Accuracy are reported to show the difference between the two statistics.}}
\label{fig:compPlot}
\end{figure*}

\paragraph{Comparison of feature extractors.}

{Here we compare our feature extraction approach against other methods that have been used for meme genre discovery.} In a second set of experiments we recomputed meme clusters using feature extractors that had previously been used in two other image cluster detection studies~\cite{zannettou2018origins,dubey2018memesequencer}. PHASH~\cite{1709989} is engineered to be robust against tiny edits, which comes from its dimensionality reduction method. The VGG Convolutional Neural Network (CNN)~\cite{simonyan2014deep} is one of the best performing pre-trained CNN models on image classification tasks. We used the version of VGG-16 trained on ImageNet~\cite{5206848}.

For this comparison experiment, we created a smaller dataset consisting of 44,612 images  selected from the original dataset. The motivation for using a smaller dataset was to allow for more human annotations per cluster and per model compared to a more demanding experiment that would require recomputing and manually checking the entire dataset for each method. The scope of this more limited, but by no means tiny, subset of the original two million images was defined by the available budget for Mechanical Turk labeling. The images were selected by randomly choosing 100 clusters that had been deemed related by the aforementioned manual annotation process. The images from the 100 clusters resulted in the new, smaller data set of 44,612 images.

We recomputed the MGD pipeline described earlier as well as the PHASH and VGG feature extractors on the reduced dataset. We set the K-means free parameter $\mathrm{K}$ to 100, resulting in the discovery of 100 distinct clusters covering the same 44,612 images for each of the algorithms. The number 100 was chosen because the smaller data set was originally divided into 100 clusters.

Using the same methodology as in the previous experiment for the entire dataset, we perform the impostor-host experiment and compare the results. Unlike MGD, which detects clusters based on the appearance and similarity of local objects, PHASH and VGG detect image clusters based on a global image similarity measure. Due to these differences, we expect that VGG and PHASH will detect many small, but very highly related clusters by grouping identical (or nearly-identical) images together; along with a few poor quality clusters, which group all of the remaining images together into very large clusters.  

As in the previous methodology, four host images and a single impostor image were randomly selected from a cluster. This process was repeated 200 times for each of the 100 clusters for each of the three algorithms. Again, annotators recruited from Amazon's Mechanical Turk service were paid 0.25 USD to annotate 25 tasks including 5 control tasks.

An unexpected effect was the generation of tiny clusters when using VGG and PHASH. Some of the clusters produced by these methods had fewer than five images, an artifact that was not observed with MGD. Thus it was possible that annotators may have been shown fewer than five images. In this case the task remained the same, but the worker simply had fewer images to choose from. Differences in the number of host images will almost certainly corrupt the methodology, so we removed annotations with fewer than four host images and one impostor image from our final results. Overall, we were able to collect 4,240 valid responses for our method: 3,374 for VGG, and 3,640 for PHASH.

\begin{table}[]
\begin{tabular}{c || c c c}
Method & MGD & VGG & PHASH \\
\hline
Number of clusters $K$ & 100 & 100 & 100 \\
Approx. run-time & 12h30m & 10h30m & 1h30m \\
Med. cluster size & 132 & 6 & 10 \\
Max. cluster size & 7,093 & 42,157 & 42,565 \\
Min. cluster size & 8 & 2 & 2 \\
Avg. Acc. & 62.42\% & 80.61\% & 38.33 \\
Normalized Avg. Acc. & 57.25\% & 26.79\% & 21.83\%  \\
Normalized Delta & 5.17\% & 53.28\% & 16.50\%
\end{tabular}
\caption{Summary statistics calculated over the smaller dataset of 44,612 images for our proposed method of local feature matching and two different global feature representations that have been used for this task.}
\label{clustermetrics}
\vspace{-2mm}
\end{table}

Comparative results are illustrated in Fig.~\ref{fig:compPlot}. These results confirm our suspicion about the distribution of accuracy ratings for the clusters. The number of images per cluster for the VGG and PHASH features is extremely skewed. Approximately 95\% of all images were placed into a single cluster for both of those features. With respect to accuracy scores, VGG features achieve an overall accuracy of 80.61\%. This is almost 20\% higher than MGD, which achieves a score of 62.42\%. While this seems to imply that VGG features are better, it is a misleading result. The picture changes significantly when the distribution of images across clusters is taken into account. With 95\% of images residing within a single cluster in the VGG case, the viewer is left with a mostly unsorted collection of images, and a set of very tiny clusters containing near-duplicate images.


In order account for the VGG and PHASH methods resulting in 95\% of images being placed in a single cluster, we also report a normalized average accuracy $\frac{1}{\mathrm{K}}\sum_{i}^{\mathrm{K}}w_i \cdot \textrm{acc}_i$, where $w_i$ is a weight defined by the percentage of images $p_i$ that appear in the $i$th cluster multiplied by a fixed scaling factor $\epsilon$, and $\textrm{acc}_i$ is the total annotator accuracy for the $i$th cluster. The normalized average accuracy along with several other statistics are shown in Table~\ref{clustermetrics}. {Notable are the deltas between accuracy and normalized average accuracy for all three approaches. VGG's performance drops over 50\% when moving to normalized average accuracy, while MGD loses only 5.17\%. Using a normalized accuracy, we find that the VGG and PHASH approaches are only slightly better than randomly generating clusters.}

\begin{figure*}[!ht]
    \centering
    \includegraphics[width=1.0\textwidth]{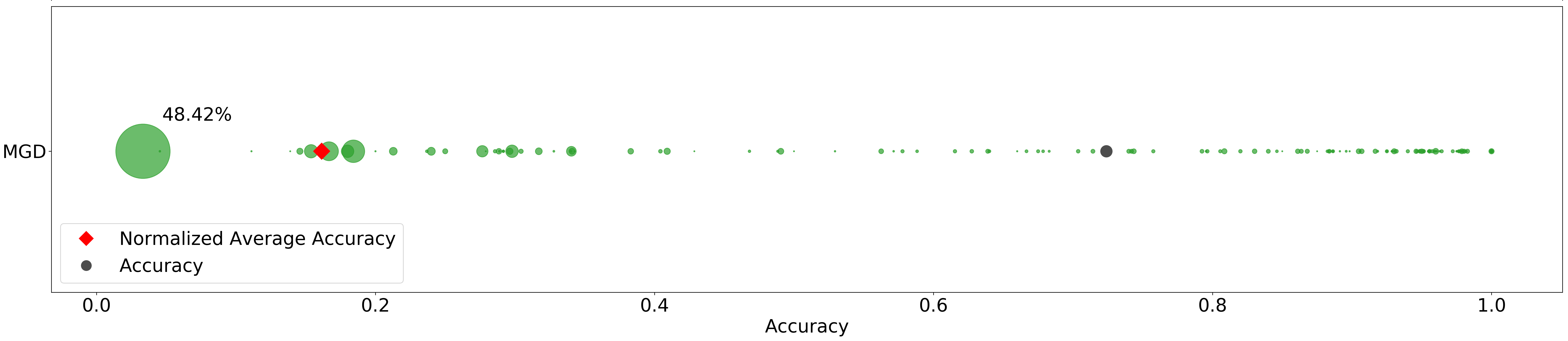}
    \vspace{-6mm}
    \caption{Shown here is the accuracy spread of the MGD clusters computed across the Reddit Photoshop Battles meme dataset~\protect\cite{moreira2018image}. In this case, normalized average accuracy is negatively impacted by having one large cluster for which annotators struggled with the associated impostor-host task. However, even with the presence of the large cluster, many good quality clusters were still assembled. Note the large number of clusters that are above 80\% accuracy.}
    \label{fig:redditTurk}
    \vspace{-3mm}
\end{figure*}

\paragraph{Generalization to other data.} In order to show that our approach generalizes to data beyond the Indonesian election, we also performed an experiment on a collection of more conventional meme images taken from the Subreddit r/photoshopbattles~\cite{moreira2018image}. This dataset was created as part of the DARPA Media Forensics program to test image provenance tools, and is far more challenging than the Indonesian election dataset. The results of this experiment are summarized in Table~\ref{redditclustermetrics} and Fig.~\ref{fig:redditTurk}. 

\begin{table}[]
\begin{tabular}{c || c c c}
Method & MGD & VGG & PHASH \\
\hline
Number of clusters $K$ & 150 & 150 & 150 \\
Approx. run-time & 4h30m & 2h & 30m \\
Med. cluster size & 13 & 1 & 1 \\
Max. cluster size & 5,042 & 10,154 & 9,629 \\
Min. cluster size & 1 & 1 & 1 \\
\end{tabular}
\caption{{Summary statistics calculated over the 
Reddit Photoshop Battles meme dataset~\protect\cite{moreira2018image}, consisting of 
10,426 images. Nearly all of the VGG and PHASH clusters contained a single image.}}
\label{redditclustermetrics}
\vspace{-6mm}
\end{table}

We can see a familiar trend in this experiment. The use of PHASH and VGG features produces a single disproportionately large cluster, while MGD more evenly distributes images across its clusters. However, a new problem with PHASH and VGG surfaced with respect to their other clusters. Of the clusters computed using VGG features, 144 of them contain a single image, or 96\%. Similarly, PHASH produced 136 clusters with a single image, or 90.67\%. Sharply contrasted to this is the small number of MGD clusters that contain a single image. Our pipeline returns 10 of these, representing just 6.67\% of the total clusters. Results of this experiment were so skewed towards small clusters for the VGG and PHASH features that it was impossible to assess their accuracy using the imposter-host task.

Not only does MGD result in a better distribution of images within clusters, but the maximum cluster size, as can be seen in Table~\ref{redditclustermetrics}, is only half the size of the maximum clusters of the other two methods. MGD achieves an accuracy of 72.39\%, calculated from 300 Turk workers performing 25 tasks as described previously. However the normalized accuracy in this case is much lower at 16.15\%. This is due to the annotators scoring significantly worse than random chance on the largest cluster in the dataset. Their accuracy on this large cluster was only 3.33\%. A potential explanation for this might be that due to the visual complexity of the dataset, they were able to find connections that weren't intended to link images. It is also possible that many related images were placed into the large cluster, even though they belong to different genres according to the meta-data of the dataset. In spite of the normalized delta being much higher in this experiment at 56.24\%, the MGD feature approach was still able to produce many more useful clusters compared to using VGG and PHASH features, and realized a high accuracy for them.

\paragraph{Sensitivity of the approach to the K-means parameter K.} To help with the selection of the K-means free parameter $\mathrm{K}$ that defines the number of clusters, we performed an experiment to study the effect $\mathrm{K}$ has on the distribution of images across the clusters. The proposed pipeline was run with all three feature types over the smaller Indonesian dataset used in the feature comparison experiments with increasingly large values of $\mathrm{K}$. 

\begin{figure}[t]
    \centering
    \includegraphics[width=0.50\textwidth]{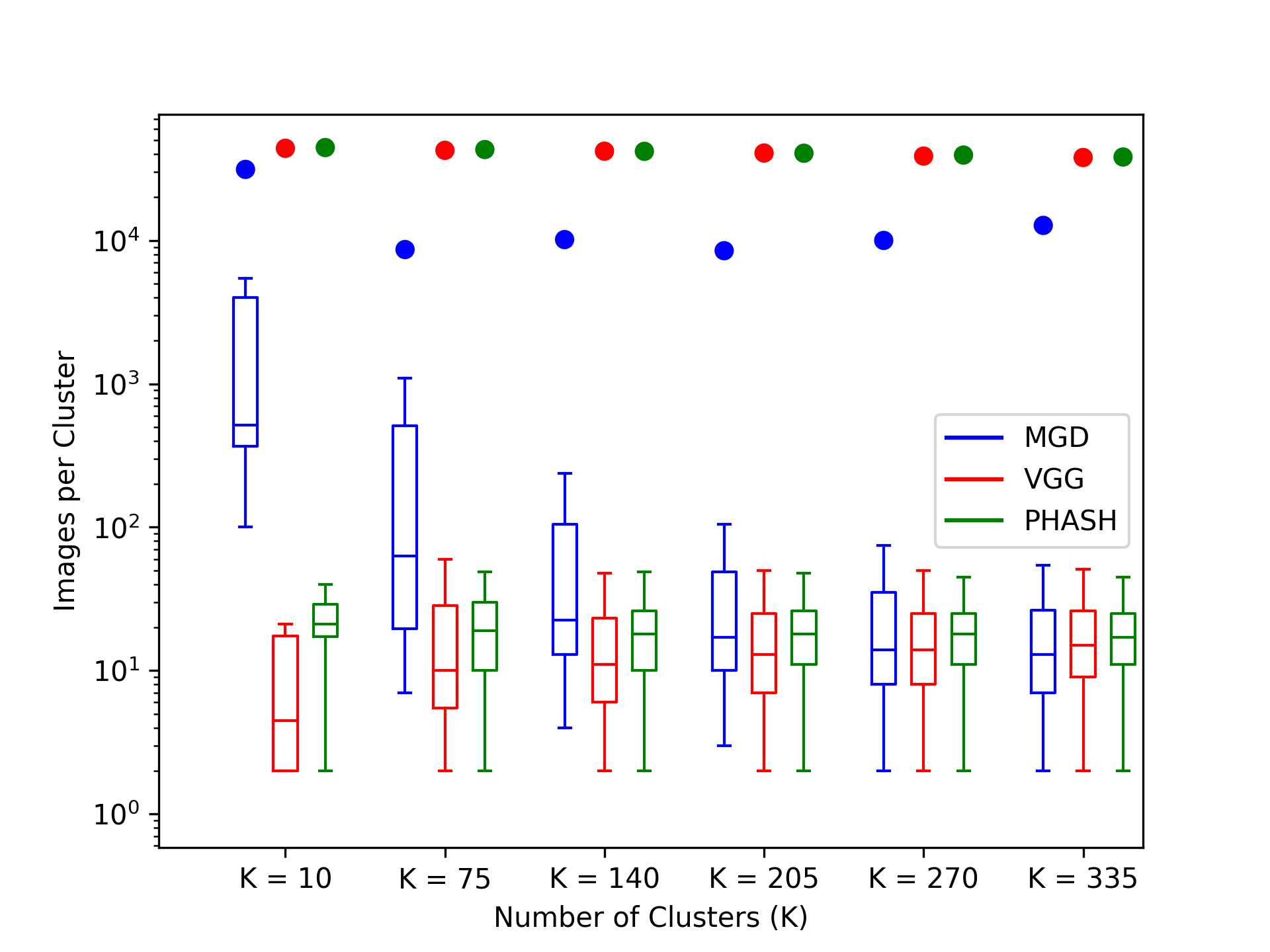}
    \vspace{-6mm}
    \caption{A comparison of images per cluster varying with the number of clusters defined by the K-means free parameter $\mathrm{K}$. These results are for the smaller Indonesian dataset used in the comparison experiment. While all three methods eventually settle on similar distributions, it is worth noting that the maxes (circles at top) for the VGG and PHASH features remain an order of magnitude above those for MGD.}
    \vspace{-3mm}
    \label{fig:kExpPlots}
\end{figure}

Fig.~\ref{fig:kExpPlots} shows that as $\mathrm{K}$ increases,  PHASH and VGG maintain a relatively stable distribution of images per cluster. The maximum cluster sizes prove very resistant to change. However as $\mathrm{K}$ moves from 10 to 140 we see a stabilization of MGD. Its distribution comes more in line with PHASH and VGG while still retaining its ability of better managing the size of the largest cluster. From this we can see that, at least for the Indonesian national election dataset, there is a range of $\mathrm{K}$ values that will produce similar results in terms of image distribution across clusters. Based on this analysis, the choice of $\mathrm{K}$ = 100 for the comparison experiments was reasonable, giving MGD more images per cluster than the other two approaches. For other datasets, a similar exploration can be performed to determine a suitable $\mathrm{K}$.

\section{Observations on the 2019 Indonesian Election}

Here we present some of our findings based on the MGD genre discovery system described in the present work. With close to two million distinct images, the Indonesian national election dataset provides insights into the technical challenges related to the clustering of meme images and other visually related content, as well as a unique perspective on a major world election. Many image datasets related to politics and social media lack such a well-defined focus, making it difficult to draw conclusions with social relevance. Because our dataset was intentionally targeted at a specific event within a limited time window, we were able to observe related activities as they unfolded. The findings discussed in this section emerged only after computing the MGD pipeline on the full dataset. Importantly, such an analysis would not have been possible if we had attempted to find relationships across all two million images by hand, especially given the sometimes small number of images in each genre.   

\begin{figure}[t]
    \centering
    \includegraphics[width=0.50\textwidth]{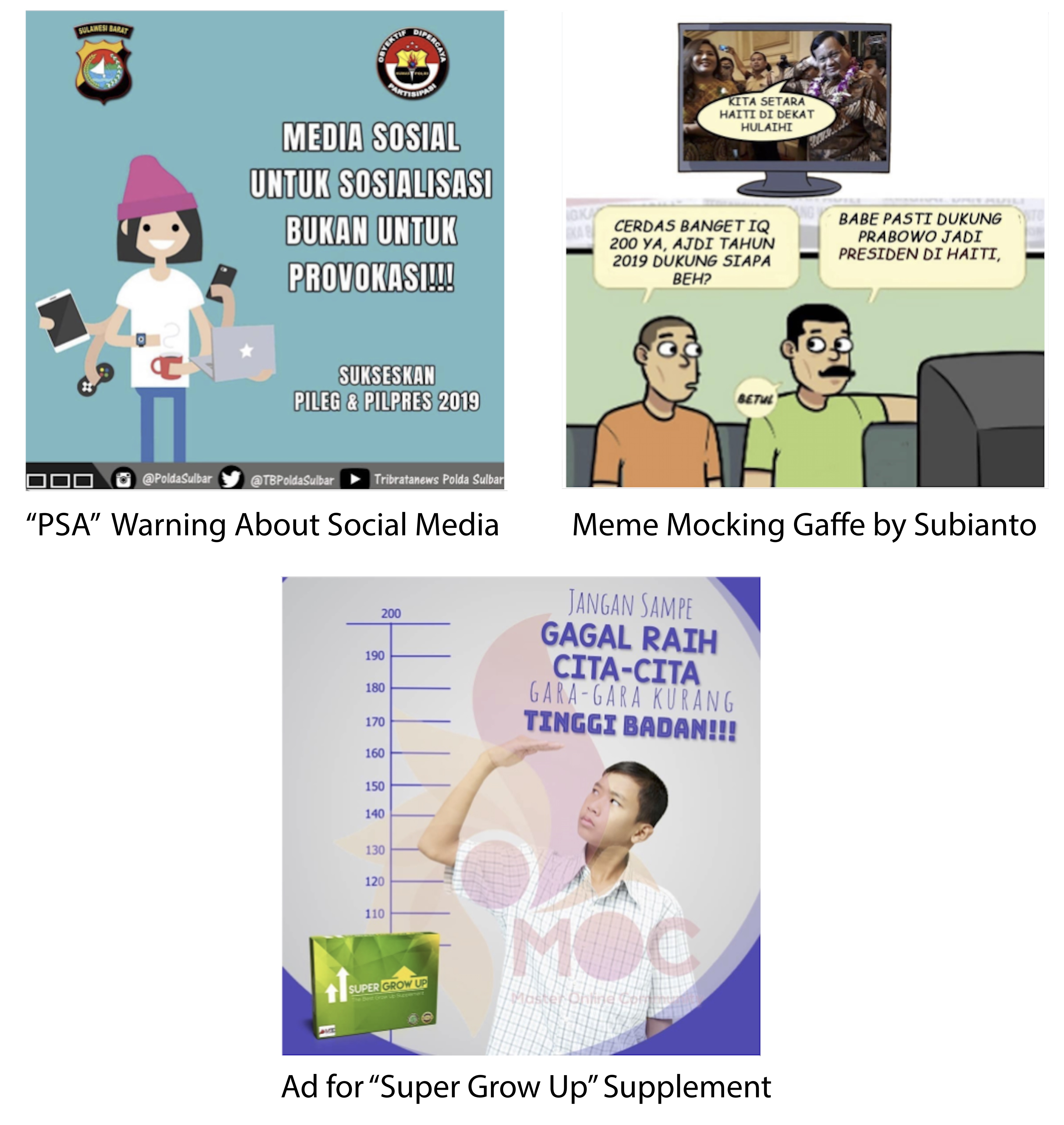}
    \vspace{-6mm}
    \caption{Examples of content posted in election-specific contexts. Top Left: A PSA type warning about misinformation released by a regional police force. Top Right: a political meme about a gaffe made by Prabowo Subianto. Bottom:  an ad for a dubious health supplement called ``Super Grow Up" claiming to increase  height.}
    \label{fig:cultureExample}
    \vspace{-3mm}
\end{figure}

As highlighted earlier, we were pleasantly surprised by the prevalence of positive messages in the dataset like the dyed finger mimicry meme. In addition to the dyed finger meme, our approach surfaced other pro-social public service announcement (PSA) messages. All of these PSAs contained some configuration of symbols for regional police organizations, which became the local image regions that were matched by MGD. These images were created to warn people to be wary of fake news and propaganda on social media during the election. For instance, in the example found on the top-left of Fig.~\ref{fig:cultureExample}, the original Bahasa text is translated to English as ``social media is for socialization not provocation.'' It is reassuring that regional authorities in the developing world are aware of the problem of disinformation and are actively working to address it. But not all of the content in the dataset contains such a positive social message.

On the campaign trail in January 2019, candidate Prabowo Subianto made the following statement regarding the status of Indonesia as a developing country: ``We, Indonesians, are on par with African impoverished countries such as Rwanda, Haiti, and small islands like Kiribati, which we don't even know where it's located''~\cite{Tempo}. This gaffe was turned into a meme by supporters of candidate Jokowi Widodo, which we only discovered after examining the clusters produced by our pipeline. The formatting of the meme places the same illustration of two men on a couch watching TV on the bottom panel, and a screenshot of what they are watching on the top panel (top-right of Fig.~\ref{fig:cultureExample}). The dialogue boxes above the men contain a conversation that mocks Subianto by riffing on his gaffe. This meme is indicative of the negative online campaigning performed by Widodo supporters.  

\begin{figure}[t]
    \centering
    \includegraphics[width=0.45\textwidth]{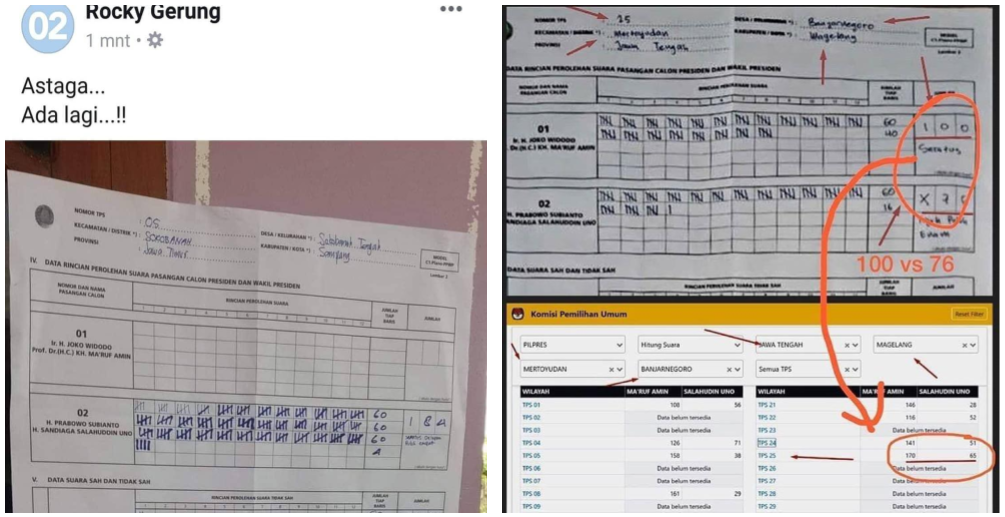}
    \caption{Two examples from a cluster showing the votes from varying districts being tallied. Notice on the right the implication that the votes that were eventually reported were not what was originally tallied.}
    \label{fig:voteTallyExample}
    \vspace{-3mm}
\end{figure}

Recall that 21.5\% of all of the images in the dataset fell into the ``Advertisement" super-genre. Much of that content was apolitical, but was collected from source accounts vigorously promoting political content on Twitter and Instagram. A curious finding was the prevalence of advertisements for a dubious health supplement called ``Super Grow Up" (bottom of Fig.~\ref{fig:cultureExample}), which promises to increase the height of adults under 40 years of  age who are taking it. This type of parasitic advertising was pioneered in the developed world by the conspiracy theory website Infowars, which promotes supplements in conjunction with its political content in the US~\cite{infowars}. Our analysis shows that this strategy is global, where significant interest in any election can translate into a large marketplace for products.

More insidious were attempts by supporters of Subianto to promote evidence of supposed voter fraud. We identified a cluster of images that was full of photos of voting tally sheets from various Indonesian voting districts. Many of these photos showed Subianto receiving more votes than Widodo in the district. After he lost the election, Subianto
alleged that there was massive, systematic fraud in the voting process~\cite{nyt-vote}. The images in Fig.~\ref{fig:voteTallyExample} are attempts to support that claim. In an era of easily accessible image manipulation tools it is difficult to tell whether or not these images have been manipulated in some way. From the massive protests that took place, we can assume that many Indonesian citizens believed them to be credible.

\section{Conclusions}

\paragraph{Summary.} This research situates itself within a growing body of work on meme analysis. The MGD pipeline shows that shortcomings in the feature extraction methods used for meme genre discovery can be addressed by tailoring the process to the localized content that is observed in different  genres. With respect to aspects of this work that other researchers will be able to make use of, we introduced a new data set of over two million images collected during the course of the 2019 Indonesian presidential election. In addition to this we are making available an open source pipeline that can, without using any meta-data, cluster images based on visual similarity with more diversity than current methods. This allows  users to more easily gain a high-level understanding of the social media landscape over time and for specific events. Both the data and code are being made available permanently through a public Google drive\footnote{Data and software are available at \url{https://bit.ly/2Rj0odI}.}.

\paragraph{Limitations of the current approach.} While we were able to match related images in a new, more refined way, the MGD approach is by no means perfect, and can benefit from further enhancements. For instance, this work aggregates the collected memes and analyzes them without any temporal context derived from meta-data, which may improve clustering accuracy. Additionally, it may be possible to fuse the output of different feature extractors. VGG features work well for near duplicate images, while our proposed local feature method is able to operate over more diverse, yet still related, images. One can envision a pipeline in which matching scores from both VGG and MGD are taken into account. Furthermore, a shortcoming of all current methods is an inability to handle images that contain objects that relate to more than one cluster. As it stands now, images will be placed into a single cluster based on the highest match score computed against a seed image. An improvement could be developing an algorithm that allows for an image to be placed in different clusters if it contains objects relevant to seed images from different genres.

Clustering a large number of images into a smaller number of clusters is very useful to human observers, but it is by no means a complete solution. Another key feature to move the technology forward would be a semantic analysis mechanism integrating visual, textual and contextual data, which can automatically provide a human-readable label to each cluster. This would further reduce the work needed for manual analysis and could allow for the application of more advanced natural language processing-based filtering methods. Such work is at the frontier of artificial intelligence, and a suitable approach does not exist yet. Even the more tractable aspects of the problem present significant challenges. For instance, reliable optical character recognition (OCR) is still an open problem in computer vision, even for very common languages such as English. Performing OCR on a relatively low resource language such as Bahasa (the most common language in Indonesia) would most likely not have yielded particularly useful information.

\paragraph{Future meme studies.} With respect to understanding memes as a form of communication, work on  GitHub communities~\cite{thomas2019dynamics} shows how groups of people working together can adopt shared linguistic libraries. Similarly, one can investigate how long it takes for a population to become ``fluent'' in a meme. Where those populations are located is likely also significant. 
Memes do not exist only in democratic countries like Indonesia. Further research is needed to understand how political memes spread in authoritarian countries that do not have democratic elections. We see that even in China, where the Internet is censored, controversial political ideas are shared online~\cite{wu2019talking}.  The approach we describe in this paper could be applied to other political events that are simultaneously unfolding in the real world and cyberspace. The Hong Kong extradition protests fit this mold.

\begin{figure}[t]
    \centering
    \includegraphics[width=0.40\textwidth]{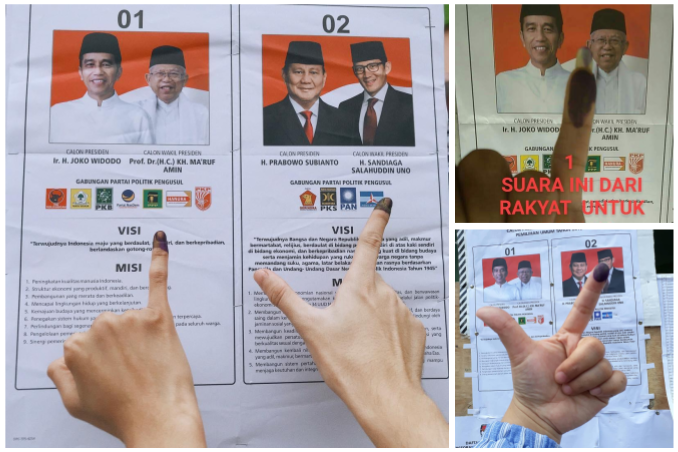}
    \caption{A crossover between the dyed finger meme and a meme that shows a number of fingers equal to the spot on the ballot of the candidate for which one was voting.}
    \label{fig:fingerSignExample}
    \vspace{-3mm}
\end{figure}

As we attempt to apply our meme discovery pipeline to collections of memes from other countries, we expect to find that memes specific to those countries would become apparent. For instance, in generated clusters from our dataset, we found hand gestures that indicated support for particular candidates (examples are shown in Fig~\ref{fig:fingerSignExample}). Supporters held up either one or two fingers, though not always the same finger(s), to indicate which spot on the ballot they were going to vote for. Such mimicry has always been commonplace in political action, and other instances where a gesture became an Internet meme abound (\textit{e.g.}, Jair Bolsonaro's signature gun gesture in the 2018 Brazilian presidential campaign). The ability of computational tools to identify such memes  represents the frontier of research in this area --- which we hope to improve on in future iterations of this work. 

\section{Acknowledgments}

This material is based on research sponsored by DARPA and the Air Force Research Laboratory (AFRL) under agreement number FA8750-16-2-0173, and the Army Research Lab under agreement number W911NF-17-1-0448, support was also given by USAID under agreement number 7200AA18CA00054. Hardware support was generously provided by the NVIDIA Corporation. Special thanks is also in order to the undergraduate data annotators: Noah Yoshida, Michael Eisemann, and Sam Battalio. Additionally Vitor Albiero was a great help with constructing several of the charts.

\bibliographystyle{aaai}
\bibliography{bibliography}

\begin{thebibliography}{}

\bibitem[\protect\citeauthoryear{Bay, Tuytelaars, and
  Van~Gool}{2006}]{bay2006surf}
Bay, H.; Tuytelaars, T.; and Van~Gool, L.
\newblock 2006.
\newblock {SURF}: Speeded up robust features.
\newblock In {\em ECCV}.

\bibitem[\protect\citeauthoryear{{BBC}}{2019}]{bbc}
{BBC}.
\newblock 2019.
\newblock Indonesia post-election protests leave six dead in {Jakarta}.

\bibitem[\protect\citeauthoryear{Beskow, Kumar, and
  Carley}{2020}]{BESKOW2020102170}
Beskow, D.~M.; Kumar, S.; and Carley, K.~M.
\newblock 2020.
\newblock The evolution of political memes: Detecting and characterizing
  {I}nternet memes with multi-modal deep learning.
\newblock {\em Information Processing \& Management} 57(2):102170.

\bibitem[\protect\citeauthoryear{Bharati \bgroup et al\mbox.\egroup
  }{2019}]{bharati2019beyond}
Bharati, A.; Moreira, D.; Brogan, J.; Hale, P.; Bowyer, K.; Flynn, P.; Rocha,
  A.; and Scheirer, W.
\newblock 2019.
\newblock Beyond pixels: Image provenance analysis leveraging metadata.
\newblock In {\em IEEE WACV}.

\bibitem[\protect\citeauthoryear{Blandfort \bgroup et al\mbox.\egroup
  }{2019}]{blandfort2019multimodal}
Blandfort, P.; Patton, D.~U.; Frey, W.~R.; Karaman, S.; Bhargava, S.; Lee,
  F.-T.; Varia, S.; Kedzie, C.; Gaskell, M.~B.; Schifanella, R.; McKeown, K.;
  and Chang, S.-F.
\newblock 2019.
\newblock Multimodal social media analysis for gang violence prevention.
\newblock In {\em AAAI ICWSM}.

\bibitem[\protect\citeauthoryear{{Brogan} \bgroup et al\mbox.\egroup
  }{2019}]{brogandynamic}
{Brogan}, J.; {Bharati}, A.; {Moreira}, D.; {Bowyer}, K.; {Flynn}, P.; {Rocha},
  A.; and {Scheirer}, W.
\newblock 2019.
\newblock Dynamic spatial verification for large-scale object-level image
  retrieval.
\newblock {\em arXiv e-prints}  arXiv:1903.10019.

\bibitem[\protect\citeauthoryear{{\v{C}}ehovin, Leonardis, and
  Kristan}{2016}]{cehovin2016visual}
{\v{C}}ehovin, L.; Leonardis, A.; and Kristan, M.
\newblock 2016.
\newblock Visual object tracking performance measures revisited.
\newblock {\em IEEE T-IP} 25(3):1261--1274.

\bibitem[\protect\citeauthoryear{{Chen} \bgroup et al\mbox.\egroup
  }{2019}]{chen2019secure}
{Chen}, Z.; {Tondi}, B.; {Li}, X.; {Ni}, R.; {Zhao}, Y.; and {Barni}, M.
\newblock 2019.
\newblock Secure detection of image manipulation by means of random feature
  selection.
\newblock {\em IEEE T-IFS} 14(9):2454--2469.

\bibitem[\protect\citeauthoryear{CNN}{2019}]{cnn}
CNN.
\newblock 2019.
\newblock Joko {Widodo} secures second term as {Indonesia's} president.

\bibitem[\protect\citeauthoryear{Coscia}{2013}]{coscia2013competition}
Coscia, M.
\newblock 2013.
\newblock Competition and success in the meme pool: a case study on
  quickmeme.com.
\newblock In {\em AAAI ICWSM}.

\bibitem[\protect\citeauthoryear{Dang \bgroup et al\mbox.\egroup
  }{2015}]{dang2015visual}
Dang, A.; Moh'd, A.; Gruzd, A.; Milios, E.; and Minghim, R.
\newblock 2015.
\newblock A visual framework for clustering memes in social media.
\newblock In {\em IEEE/ACM ASONAM}.

\bibitem[\protect\citeauthoryear{de Oliveira \bgroup et al\mbox.\egroup
  }{2016}]{oliveira2016multiple}
de~Oliveira, A.; Ferrara, P.; De~Rosa, A.; Piva, A.; Barni, M.; Goldenstein,
  S.; Dias, Z.; and Rocha, A.
\newblock 2016.
\newblock Multiple parenting phylogeny relationships in digital images.
\newblock {\em IEEE T-IFS} 11(2):328--343.

\bibitem[\protect\citeauthoryear{{Deng} \bgroup et al\mbox.\egroup
  }{2009}]{5206848}
{Deng}, J.; {Dong}, W.; {Socher}, R.; {Li}, L.; {Kai Li}; and {Li Fei-Fei}.
\newblock 2009.
\newblock {ImageNet}: A large-scale hierarchical image database.
\newblock In {\em IEEE CVPR}.

\bibitem[\protect\citeauthoryear{Dewan \bgroup et al\mbox.\egroup
  }{2017}]{dewan2017towards}
Dewan, P.; Suri, A.; Bharadhwaj, V.; Mithal, A.; and Kumaraguru, P.
\newblock 2017.
\newblock Towards understanding crisis events on online social networks through
  pictures.
\newblock In {\em IEEE/ACM ASONAM}.

\bibitem[\protect\citeauthoryear{Dias, Goldenstein, and
  Rocha}{2013}]{dias2013large}
Dias, Z.; Goldenstein, S.; and Rocha, A.
\newblock 2013.
\newblock Large-scale image phylogeny: Tracing image ancestral relationships.
\newblock {\em IEEE Multimedia} 20(3):58--70.

\bibitem[\protect\citeauthoryear{Dubey \bgroup et al\mbox.\egroup
  }{2018}]{dubey2018memesequencer}
Dubey, A.; Moro, E.; Cebrian, M.; and Rahwan, I.
\newblock 2018.
\newblock Memesequencer: Sparse matching for embedding image macros.
\newblock In {\em IW3C2 Web Conference}.

\bibitem[\protect\citeauthoryear{Farid}{2016}]{farid2016photo}
Farid, H.
\newblock 2016.
\newblock {\em Photo Forensics}.
\newblock MIT Press.

\bibitem[\protect\citeauthoryear{Gal, Shifman, and Kampf}{2016}]{gal2016gets}
Gal, N.; Shifman, L.; and Kampf, Z.
\newblock 2016.
\newblock ``{It} gets better'': Internet memes and the construction of
  collective identity.
\newblock {\em New Media \& Society} 18(8):1698--1714.

\bibitem[\protect\citeauthoryear{Ge \bgroup et al\mbox.\egroup
  }{2013}]{ge2013optimized}
Ge, T.; He, K.; Ke, Q.; and Sun, J.
\newblock 2013.
\newblock Optimized product quantization for approximate nearest neighbor
  search.
\newblock In {\em IEEE CVPR}.

\bibitem[\protect\citeauthoryear{Huh \bgroup et al\mbox.\egroup
  }{2018}]{huh2018fighting}
Huh, M.; Liu, A.; Owens, A.; and Efros, A.~A.
\newblock 2018.
\newblock Fighting fake news: Image splice detection via learned
  self-consistency.
\newblock In {\em ECCV}.

\bibitem[\protect\citeauthoryear{Jinda-Apiraksa, Vonikakis, and
  Winkler}{2013}]{winkler2013nd}
Jinda-Apiraksa, A.; Vonikakis, V.; and Winkler, S.
\newblock 2013.
\newblock {California-ND: An annotated dataset for near-duplicate detection in
  personal photo collections}.
\newblock In {\em IEEE QoMEX}.

\bibitem[\protect\citeauthoryear{Lowry \bgroup et al\mbox.\egroup
  }{2016}]{lowry2016visual}
Lowry, S.; S{\"u}nderhauf, N.; Newman, P.; Leonard, J.~J.; Cox, D.; Corke, P.;
  and Milford, M.~J.
\newblock 2016.
\newblock Visual place recognition: A survey.
\newblock {\em IEEE T-RO} 32(1):1--19.

\bibitem[\protect\citeauthoryear{{Monga} and {Evans}}{2006}]{1709989}
{Monga}, V., and {Evans}, B.~L.
\newblock 2006.
\newblock Perceptual image hashing via feature points: Performance evaluation
  and tradeoffs.
\newblock {\em IEEE T-IP} 15(11):3452--3465.

\bibitem[\protect\citeauthoryear{Moreira \bgroup et al\mbox.\egroup
  }{2018}]{moreira2018image}
Moreira, D.; Bharati, A.; Brogan, J.; Pinto, A.; Parowski, M.; Bowyer, K.~W.;
  Flynn, P.~J.; Rocha, A.; and Scheirer, W.~J.
\newblock 2018.
\newblock Image provenance analysis at scale.
\newblock {\em IEEE T-IP} 27(12):6109--6123.

\bibitem[\protect\citeauthoryear{Nissenbaum and
  Shifman}{2017}]{nissenbaum2017internet}
Nissenbaum, A., and Shifman, L.
\newblock 2017.
\newblock Internet memes as contested cultural capital: The case of 4chan's /b/
  board.
\newblock {\em New Media \& Society} 19(4):483--501.

\bibitem[\protect\citeauthoryear{Rumata and
  Sastrosubroto}{2018}]{rumata2018net}
Rumata, V.~M., and Sastrosubroto, A.~S.
\newblock 2018.
\newblock Net-attack 2.0: Digital post-truth and its regulatory challenges in
  {Indonesia}.
\newblock In {\em ICCSR}.

\bibitem[\protect\citeauthoryear{Russakovsky \bgroup et al\mbox.\egroup
  }{2015}]{russakovsky2015imagenet}
Russakovsky, O.; Deng, J.; Su, H.; Krause, J.; Satheesh, S.; Ma, S.; Huang, Z.;
  Karpathy, A.; Khosla, A.; Bernstein, M.; et~al.
\newblock 2015.
\newblock {ImageNet} large scale visual recognition challenge.
\newblock {\em IJCV} 115(3):211--252.

\bibitem[\protect\citeauthoryear{Seiffert-Brockmann, Diehl, and
  Dobusch}{2018}]{seiffert2018memes}
Seiffert-Brockmann, J.; Diehl, T.; and Dobusch, L.
\newblock 2018.
\newblock Memes as games: The evolution of a digital discourse online.
\newblock {\em New Media \& Society} 20(8):2862--2879.

\bibitem[\protect\citeauthoryear{Shifman}{2012}]{shifman2011anatomy}
Shifman, L.
\newblock 2012.
\newblock An anatomy of a {YouTube} meme.
\newblock {\em New Media \& Society} 14(2):187--203.

\bibitem[\protect\citeauthoryear{Shifman}{2013}]{shifman2014memes}
Shifman, L.
\newblock 2013.
\newblock {\em Memes in Digital Culture}.
\newblock The MIT Press.

\bibitem[\protect\citeauthoryear{Simonyan and
  Zisserman}{2014}]{simonyan2014deep}
Simonyan, K., and Zisserman, A.
\newblock 2014.
\newblock Very deep convolutional networks for large-scale image recognition.
\newblock {\em arXiv preprint arXiv:1409.1556}.

\bibitem[\protect\citeauthoryear{Stella and Shi}{2003}]{stella2003multiclass}
Stella, X.~Y., and Shi, J.
\newblock 2003.
\newblock Multiclass spectral clustering.
\newblock In {\em IEEE ICCV}.

\bibitem[\protect\citeauthoryear{TEMPO.CO}{2018}]{Tempo}
TEMPO.CO.
\newblock 2018.
\newblock Prabowo likens {Indonesia's} economy to {African} country {Haiti}.

\bibitem[\protect\citeauthoryear{{The New York Times}}{2019}]{nyt-vote}
{The New York Times}.
\newblock 2019.
\newblock Indonesia court rejects presidential candidate’s voting fraud
  claims.

\bibitem[\protect\citeauthoryear{{The Washington Post}}{2018}]{infowars}
{The Washington Post}.
\newblock 2018.
\newblock As {Alex Jones} rails against ‘big tech,’ his {Infowars} stores
  still thrive online.

\bibitem[\protect\citeauthoryear{{The World Bank}}{2019}]{WorldBank}
{The World Bank}.
\newblock 2019.
\newblock The {World Bank} in {Indonesia}.

\bibitem[\protect\citeauthoryear{Thomas, Krohn, and
  Weninger}{2019}]{thomas2019dynamics}
Thomas, P.~B.; Krohn, R.; and Weninger, T.
\newblock 2019.
\newblock Dynamics of team library adoptions: An exploration of {GitHub} commit
  logs.
\newblock {\em arXiv preprint arXiv:1907.04527}.

\bibitem[\protect\citeauthoryear{{Wall Street Journal}}{2014}]{wsj}
{Wall Street Journal}.
\newblock 2014.
\newblock Indonesian voters dip their fingers and pose.

\bibitem[\protect\citeauthoryear{Weninger, Bisk, and
  Han}{2012}]{10.1145/2396761.2396843}
Weninger, T.; Bisk, Y.; and Han, J.
\newblock 2012.
\newblock Document-topic hierarchies from document graphs.
\newblock In {\em CIKM},  635–644.

\bibitem[\protect\citeauthoryear{Wu and Mai}{2019}]{wu2019talking}
Wu, S., and Mai, B.
\newblock 2019.
\newblock Talking about and beyond censorship: Mapping topic clusters in the
  {Chinese} {Twitter} sphere.
\newblock {\em International Journal of Communication} 13:23.

\bibitem[\protect\citeauthoryear{Zannettou \bgroup et al\mbox.\egroup
  }{2018}]{zannettou2018origins}
Zannettou, S.; Caulfield, T.; Blackburn, J.; De~Cristofaro, E.; Sirivianos, M.;
  Stringhini, G.; and Suarez-Tangil, G.
\newblock 2018.
\newblock On the origins of memes by means of fringe web communities.
\newblock In {\em ACM Internet Measurement Conference},  188--202.

\bibitem[\protect\citeauthoryear{Zhou \bgroup et al\mbox.\egroup
  }{2018}]{zhou2017places}
Zhou, B.; Lapedriza, A.; Khosla, A.; Oliva, A.; and Torralba, A.
\newblock 2018.
\newblock Places: A 10 million image database for scene recognition.
\newblock {\em IEEE T-PAMI} 40(6):1452--1464.

\end{thebibliography}


\section{Appendix: Data Sources}

Some accounts listed here have since been banned.

\begin{table}[h]
\begin{center}
\begin{tabular}{ c c }
 Twitter Hashtags & Twitter Users  \\ 
 \hline
  2019gantipresiden & maspiyuuu \\
 jokowiudahkasihbukti & MCAOps\\ 
 prabowosandibawasolusi & jokowi  \\
 KoalisiPraBohong & \\
 prabowotakuttesngaji &  \\
 SandiwaraUno & \\
 GaHoaxGaMakan &  \\
 SemburanDusta02 &  \\
 menujuindonesiamaju & \\
 jokowitakutpaparkanmisivisi & \\
 prabowoindonesiamenang & \\
\end{tabular}
\end{center}
\end{table}


\begin{table}[h]
\begin{center}
\begin{tabular}{ c c}
Instagram Hashtags & Instagram Users \\ 
 \hline
   01optimisindonesiamaju & jihad\_konstitusi \\  
  salamwaras & indonesiavoice\_ \\ 
  2019jokowiamin & dhidysatriady \\
 jokowilagi & kangrepost21 \\
  tangkapaminrais &  \_velvetsky02\_ \\
 dagelan & \\
  jokowinelection & \\
 tangkapprabowo & \\
 tangkapsandi & \\
  penjarakanaminrais & \\
  tangkaptitieksuharto &\\
 penjarakanprabowo & \\
 penjarakanprabowosandi & \\
 vivatnipolri & \\
 pilpres2019 & \\
\end{tabular}
\end{center}
\end{table}

\end{document}